\documentclass{IEEEcsmag}

\usepackage[colorlinks,urlcolor=blue,linkcolor=blue,citecolor=blue]{hyperref}
\expandafter\def\expandafter\UrlBreaks\expandafter{\UrlBreaks\do\/\do\*\do\-\do\~\do\'\do\"\do\-}
\usepackage{upmath,color}

\usepackage{amsmath,amsfonts}
\usepackage{algorithmic}
\usepackage[caption=false,font=normalsize,labelfont=sf,textfont=sf]{subfig}
\usepackage{textcomp}
\usepackage{stfloats}
\usepackage{url}
\usepackage{verbatim}
\usepackage{graphicx}
\def\BibTeX{{\rm B\kern-.05em{\sc i\kern-.025em b}\kern-.08em
    T\kern-.1667em\lower.7ex\hbox{E}\kern-.125emX}}
\usepackage{balance}
\usepackage{booktabs}
\usepackage{array}
\usepackage{cleveref}

\usepackage{makecell}
\usepackage{array}
\usepackage{tikz}
\usetikzlibrary{trees}
\usepackage{tabularx,colortbl}
\usepackage{booktabs} 
\usepackage{multirow}
\usepackage{float}
\usepackage{placeins}
\usepackage{caption}
\usepackage{chngcntr}

\jvol{XX}
\jnum{XX}
\jmonth{May}
\jname{Publication Name}
\jtitle{Assisting Research Proposal Writing with LLMs}
\pubyear{2025}

\begin{document}

\sptitle{Assisting Research Proposal Writing with LLMs}

\title{Assisting Research Proposal Writing with Large Language Models: Evaluation and Refinement}

\author{Jing Ren}
\affil{University of Technology Sydney, Sydney, NSW, 2007, Australia}

\author{Weiqi Wang}
\affil{University of Technology Sydney, Sydney, NSW, 2007, Australia}

\markboth{THEME/FEATURE/DEPARTMENT}{THEME/FEATURE/DEPARTMENT}

\begin{abstract}\looseness-1Large language models (LLMs) like ChatGPT are increasingly used in academic writing, yet issues such as incorrect or fabricated references raise ethical concerns. Moreover, current content quality evaluations often rely on subjective human judgment, which is labor-intensive and lacks objectivity, potentially compromising the consistency and reliability. In this study, to provide a quantitative evaluation and enhance research proposal writing capabilities of LLMs, we propose two key evaluation metrics--content quality and reference validity--and an iterative prompting method based on the scores derived from these two metrics. Our extensive experiments show that the proposed metrics provide an objective, quantitative framework for assessing ChatGPT’s writing performance. Additionally, iterative prompting significantly enhances content quality while reducing reference inaccuracies and fabrications, addressing critical ethical challenges in academic contexts.


\end{abstract}

\maketitle

\chapteri{L}arge language models (LLMs) like ChatGPT offers writing support in areas such as grammar, vocabulary, and style \cite{song2023enhancing}; however, the misuse of AI tools in academic writing can threaten scientific integrity. Ethical concerns, including issues of falsified references \cite{awosanya2024utility} and the spread of misinformation due to unchecked AI-generated outputs \cite{buruk2023academic} may raise worries that AI could contribute to the production of low-quality manuscripts containing false information as it becomes more widely adopted in scientific research writing~\cite{kacena2024use}. For a research proposal, every aspect must be carefully planned, with researchers considering the implications and potential challenges \cite{bell2018ebook}. Additionally, a well-rounded research proposal requires the integration of various elements to ensure clarity, focus, and feasibility \cite{cdi_gale_businessinsightsgauss_A795341205}. Each cited reference provides detailed insights into the key components of proposal writing, making it essential that all sources are properly cited in the bibliography. In light of the rigorous standards in research proposal writing, questions remain as to whether AI can generate credible and publishable full-length scientific research proposals.

Current research provides limited attention to LLM-generated research proposals, lacks quantifiable evaluation of content quality, and offers few solutions for enhancing writing capabilities. Previous research such as \cite{awosanya2024utility,kacena2024use} mainly focus on review paper writing but lack of research proposal generation. Besides, current evaluations of LLMs' writing capabilities rely largely on subjective human judgment, lacking an objective methodology for assessing the quality of generated content \cite{bucol2024exploring}. Moreover, studies have identified inaccuracies and fabrications in references generated by ChatGPT in academic writing~\cite{awosanya2024utility}. Despite these concerns, effective solutions to mitigate these issues remain limited.

\textbf{Research Question.} Based on the existing research, {we try to answer the following research questions in this paper:} 

\textit{RQ1: How can writing capabilities of LLMs be evaluated through a rigorous, scientific, and quantifiable approach in research proposal writing?}


\textit{RQ2: How can the writing capabilities of LLMs be effectively enhanced in research proposal writing?}

\begin{figure*}[t]
    \centering
    \includegraphics[width=0.94\textwidth]{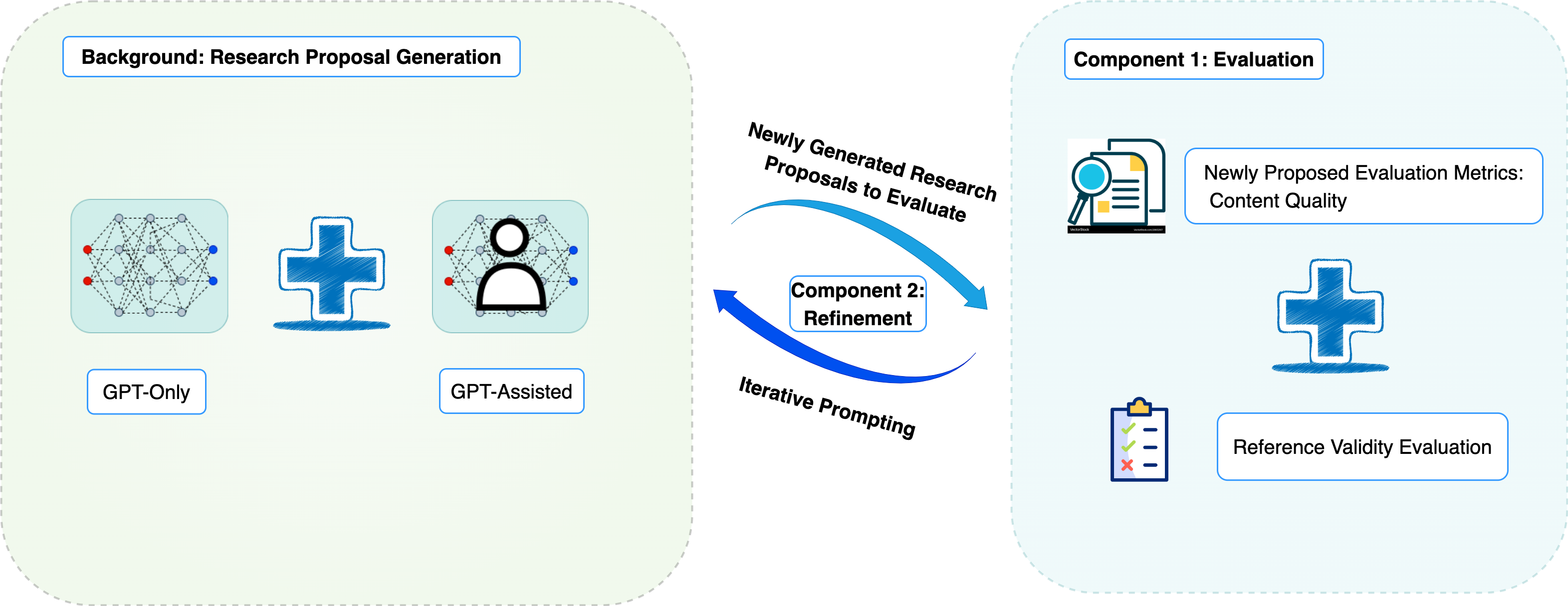} 
    \caption{Process overview: evaluating and enhancing LLM writing capabilities from proposal generation to assessment and refinement}
    \label{overview_process}
\end{figure*}

\textbf{Our Work.} This study aims to evaluate LLMs' effectiveness and potential in supporting and enhancing the research proposal development process. An overview of our approach is presented in \Cref{overview_process}. Through this integrated framework—encompassing proposal generation, assessment, and refinement—LLM-based writing can be more effectively aligned with the rigorous demands of scientific research proposal writing.


To address RQ1, we introduced two evaluation metrics—content quality and reference validity. Unlike traditional evaluation methods that rely heavily on human judgment and often overlook the quality of written content, our approach provides an efficient, objective, and quantitative framework for assessing LLMs' capabilities in research proposal writing. Once the research proposals were generated, they were evaluated using an AI grading system to assess content quality, while reference inaccuracies and fabrications were examined separately.

To address RQ2, we proposed an iterative prompting method to improve the content quality of research proposals and reduce errors and fabrications in references. This approach involves continuous adjustments and repeated prompts, where scores from the AI grading system are fed back into LLM as targeted feedback to guide further refinement. We also investigated the conditions under which LLMs can effectively minimize reference inaccuracies and fabrications, ultimately producing more stable and reliable outputs. This method establishes a positive, iterative cycle in which LLMs' performance is progressively enhanced—from initial proposal generation to evaluation and continuous improvement through iterative prompting.

We conduct extensive experiments through dual-metric evaluation and iterative prompting approach. The results demonstrate that our proposed metrics offer an efficient, objective, and quantitative framework for assessing LLMs' writing capabilities. Moreover, the experimental findings confirm that the iterative prompting method effectively improves content quality and substantially reduces errors and fabrications in reference generation.

Our contributions can be summarized as belows:
\begin{itemize}
     \item {We introduce a key evaluation metric—content quality, assessed through an AI grading system—alongside reference validity, to provide a clear, objective, consistent, and quantitative assessment of LLMs' writing capabilities. This framework effectively addresses the limitations of traditional human evaluations, such as inaccuracy, inconsistency, and subjectivity.}
     \item {We propose an iterative prompting method guided by evaluation metrics to continuously improve LLMs' writing capabilities—from proposal generation to assessment and refinement—ensuring that the generated proposals more effectively align with rigorous academic standards.}
    \item {We conduct extensive experiments, and the results demonstrate the effectiveness of the iterative prompting method in enhancing the content quality of research proposals and mitigating inaccuracies and fabrications in referencing.}
   
\end{itemize}


\section{RELATED WORK}
\label{Related works}

While numerous studies explore ChatGPT's use in academic writing, few specifically address its role in crafting academic proposals. AI tools that incorporate grammar and style checking are already widespread, but ChatGPT-4 takes this a step further by suggesting not just grammatical corrections but also stylistic adjustments that align with formal academic writing \cite{du2024exploring}. In academic writing, based on current studies, ChatGPT can be particularly beneficial in various stages of the writing and research process, offering support in areas such as brainstorming~\cite{lavrivc2023brainstorming}, drafting \cite{picciano2024graduate}, revising \cite{alyasiri2024chatgpt}, and even performing linguistic analysis \cite{mizumoto2023exploring}. 

\FloatBarrier
\begin{table*}[t!]
    \scriptsize
    \centering
    \caption{The difference between existing methods}
    \label{difference}
    \resizebox{\linewidth}{!}{
        \setlength\tabcolsep{1.pt}
        \begin{tabular}{p{3.5cm}|p{5cm} p{6.5cm}}
            \toprule[0.8pt]
            \textbf{Aspects} & \textbf{Previous Research} & \textbf{This Research} \\
            \midrule
            \addlinespace
            \midrule
            \textbf{Writing Evaluation Metrics} & 
            Linguistic Features \cite{lee2022coauthor,mizumoto2023exploring}, Time Consumption, Similarity, \textbf{Reference Validity~\cite{kacena2024use}} & $\to$ \textbf{Content Quality, References Validity}\\
            \addlinespace
            \midrule
            \textbf{Content Evaluation Methods} & 
            Human Evaluation \cite{kacena2024use} &  $\to$ \text{
            \textbf{AI Evaluation} } \\
            \addlinespace
            \midrule
              \textbf{Approaches to Enhancing GPT's Writing Capabilities} & 
            None &  $\to$ \textbf{
            Iterative Prompting Method} \\
            \bottomrule[0.8pt]
        \end{tabular}
    }
\end{table*}
\FloatBarrier


There is a lack of quantitative evaluation and effective enhancement strategies for GPT's writing capabilities. Assessments of GPT-generated writing quality are often based on human comparisons that consider aspects such as reference accuracy and similarity~\cite{kacena2024use}, grammatical correctness, and vocabulary diversity \cite{lee2022coauthor}. This highlights the need for a comprehensive, objective, and quantitative assessment of ChatGPT’s writing capabilities. Such an evaluation should address key aspects including grammar, fluency, clarity, relevance, organization, style, source formatting, and adherence to academic conventions. Furthermore, existing research identifies several challenges and limitations of ChatGPT in academic contexts, including issues related to accuracy and reliability \cite{lo2023impact}, ethical concerns and biases \cite{bolukbasi2016man}, the ongoing need for model improvement \cite{chukwuere2024today}, risks of plagiarism \cite{awosanya2024utility}, and repetitive language use \cite{cotton2024chatting}. Despite the recognition of these challenges, effective solutions are still limited.

ChatGPT-4o is the latest version of OpenAI’s generative language model, built on the powerful GPT-4 architecture, designed to assist users in generating human-like text based on given prompts. Our study uses ChatGPT-4o for the generation of research proposals with the aim of improving the quality of writing. The proposed dual-metric framework—comprising content quality and reference validity—provides a clear, objective, and quantitative approach to assessing ChatGPT’s writing capabilities. Content quality is evaluated using an AI-based grading system, offering a more efficient and time-saving alternative to traditional human evaluation. Both content quality and reference validity are further improved through an iterative prompting method. Through rigorous experiments and analysis, we demonstrate the effectiveness of our quantitative evaluation metrics and iterative prompting approach. The following section outlines the key differences between our research and previous studies (see \Cref{difference}).

\section{PROBLEM STATEMENT}
\label{PS}


A research proposal is a structured document that outlines a planned research project, serving as a roadmap for the study. It specifies the research problem, objectives, methodology, and anticipated outcomes \cite{cdi_gale_businessinsightsgauss_A795341205}. An effective research proposal brings together key elements to promote clarity, coherence, and practicality, with accurate citation and proper referencing serving as critical components of scholarly writing. 

Our objective in utilizing LLMs for research proposal drafting is to guide it in producing high-quality, academically sound proposals. By refining LLM-generated outputs, we aim to make them a valuable foundation for researchers. An effective LLM-assisted research proposal writing framework should meet the following key requirements.

\begin{itemize}
     \item {\textit{Objective and quantitative evaluation metrics.} These metrics assess LLMs' writing capabilities by evaluating content quality—alongside reference validity—across multiple dimensions, including grammar, fluency, clarity, relevance, organization, style, source accuracy, and adherence to academic standards.}
    \item {\textit{A strategy to enhance LLMs' writing capabilities.} This design works as a mechanism for continual improvement of proposals generated by LLMs. To be specific, this method can iteratively improve content quality and reduces inaccuracies and fabrications in reference generation.}
   
\end{itemize}

\section{METHODOLOGY}

\label{Methods}

\subsection{Overview}



This research employs a two-metric evaluation and an iterative prompting method to investigate the effectiveness and potential of LLMs in research proposal generation. Based on ChatGPT-4o, the two-metric evaluation assesses LLMs’ writing capabilities by measuring both content quality and reference validity. The iterative prompting method is designed to enhance LLMs’ writing performance by improving content quality and reducing reference inaccuracies and fabrications.

\subsection{Metrics Assessing GPT's Writing Capability}

In this research, we propose two evaluation metrics to provide a comprehensive, objective, and quantitative assessment of GPT’s writing performance: content quality and reference validity. This framework evaluates both the written content—covering grammar, fluency, clarity, relevance, organization, style, source formatting, and adherence to academic conventions—and the accuracy of references, including formatting consistency, correctness, and the presence of fabricated citations.

Once each research proposal was generated, its content quality and reference validity were independently assessed. Content quality was evaluated using an AI-based grading process, with scores averaged across three professional platforms: Study Fetch, QuillBot \cite{gurbuz2024impact}, and Grammarly \cite{ding2024automated}. These platforms were selected for their use of advanced AI algorithms, continuous model updates, alignment with educational standards, and transparent feedback mechanisms. 

Collectively, Study Fetch, QuillBot \cite{gurbuz2024impact}, and Grammarly \cite{ding2024automated} offer an efficient and reliable method for evaluating academic writing, significantly reducing the time and effort required for manual review. Study Fetch evaluates proposals across multiple academic dimensions, including the research problem statement, literature review, theoretical framework, research objectives/questions, methodology, expected contributions, organization and coherence, writing quality, and references. QuillBot focuses on grammar, fluency, clarity, and engagement, while Grammarly provides assessments related to word choice and overall readability

Reference validity was assessed through a manual fact-checking process conducted by the first two authors. Each citation generated by GPT was verified for accuracy, including details such as author names, publication titles, publishers, and page numbers. This step ensured the credibility and academic suitability of the proposals, preventing the inclusion of fabricated or misleading references.

\subsection{Iterative Prompting Method}

LLM-generated research proposals may still contain falsified or fabricated information, and their content quality remains an area for further improvement. The iterative prompting method is applied based on the results of two metrics evaluation with the aim of enhancing content quality and ensuring reference validity. To enhance the content quality of research proposals, the scores obtained from the AI scoring systems are fed back to GPT as targeted feedback, focusing on specific writing weaknesses such as clarity and fluency. Based on this feedback, GPT revises the proposal with the goal of improving those identified areas. The revised proposal is then re-evaluated using the same AI scoring systems. If certain aspects still do not meet the desired standards, the detailed scoring results are once again used as feedback for further refinement. This process continues iteratively until the writing strategy produces a research proposal that meets quality expectations. 


Falsified references and fabrications \cite{awosanya2024utility} pose a threat to the integrity of research. To mitigate this, we propose an itrative prompting method to minimize inaccuracies and fabrications while enhancing the correctness of references generated by LLMs. After each research proposal was generated, the authors provided a reference guide for review. This iterative process was applied to refine GPT's outputs by correcting errors in format, author, title, year, journal, volume, page number, and publisher, identifying fabricated references, and ensuring consistency between in-text citations and the reference list. The correction process was repeated multiple times to achieve stable and improved performance (see Appendix~\ref{Appendix: Improving Reference Accuracy} for details).

\section{PERFORMANCE EVALUATION}
\label{PE}

\subsection{Experimental Settings}

\textbf{Writing Strategies.} This experiment uses a ChatGPT-4o model as a writing strategy for generating research proposals. To be specific, two writing strategies for research proposal generation are designed: (1) \textbf{GPT-only}, where proposals are generated solely by standard ChatGPT-4o without human intervention; (2) \textbf{GPT-assisted}, where ChatGPT-4o generates proposals using human-provided references. The written details are shown below.

{GPT-Only Writing Strategy:} To examine GPT’s standalone writing performance and compare the results against the other writing strategy, we generated a research proposal written entirely by GPT without any human intervention. This was accomplished by submitting a single query modified from \cite{kacena2024use} within the context of academic writing, specifying the writer's identity, writing level, word count, topic, requirements for each section, and reference formatting. The writing query can be found in \Cref{appendix:gpt_only}.



{GPT-Assisted Writing Strategy:} To enhance the quality of GPT's proposal writing, the authors provided a selection of highly relevant references for the model to reference. The reference provision process and proposal generation queries were formulated based on AI-assisted writing guidelines outlined in \cite{kacena2024use}. Queries are presented in \Cref{appendix:gpt_assisted}.

\textbf{Proposal Writing Topics.} Proposals related to three distinct topics in the education field are written: (1) \textit{Factors Affecting the Agency of University Students (Student Agency)}; (2) \textit{Factors Affecting the Agency of University Teachers (Teacher Agency)}; and (3) \textit{The Identity of University Teachers (Teacher Identity)}. The topics were selected for two main reasons. First, they are widely recognized themes in education research \cite{etelapelto2013agency}, representing dynamic and evolving fields with ongoing knowledge gaps, making them suitable for analyzing and interpreting current research data. Second, as specialists in \textit{Agency} and \textit{Identity}, we are well-positioned to critically evaluate the validity and relevance of research inquiries within these domains.

\textbf{Metrics.} This research utilizes a two-metric evaluation framework—content quality and reference validity—to assess GPT’s writing capabilities. Content quality was evaluated through a grading process, with scores averaged across three grading platforms: Study Fetch, QuillBot \cite{gurbuz2024impact}, and Grammarly~\cite{ding2024automated}. Each citation generated by ChatGPT was verified for accuracy and any fabricated information.

\textbf{Iterative Prompting Methods.} After each research proposal was generated, the content quality scores from the AI grading system were used as feedback to target specific areas needing improvement. The revised proposal was then re-evaluated using the same scoring system to assess content quality. This evaluation and refinement process was conducted iteratively until the proposal met the desired quality standards. Additionally, the authors provided GPT with a reference guide to review, learn from, and correct inaccuracies, while identifying and eliminating fabricated references. This correction process was repeated over multiple rounds to ensure consistent improvement and reliable performance.

\begin{table*}[!t]
\centering
\small
\caption{Evaluation of content quality: grading software performance and topic-wise average scores (GPT-only vs. GPT-assisted)}
\resizebox{\linewidth}{!}{
\begin{tabular}{lcccccc|cc}
\toprule
\multirow{2}{*}{Writing Topics} & \multicolumn{2}{c}{Study Fetch} & \multicolumn{2}{c}{QuillBot} & \multicolumn{2}{c|}{Grammarly} & \multicolumn{2}{c}{Average Score} \\
\cmidrule(r){2-7} \cmidrule(l){8-9}
& GPT-Only & GPT-Assisted & GPT-Only & GPT-Assisted & GPT-Only & GPT-Assisted & GPT-Only & GPT-Assisted \\ 
\midrule
Student Agency   & 71 & 79 & 81 & 87 & 92 & 94 & 81.33 & 86.67 \\
Teacher Agency   & 81 & 82 & 82 & 83 & 92 & 95 & 85.00 & 86.67 \\
Teacher Identity & 82 & 72 & 81 & 81 & 91 & 92 & 84.67 & 81.67 \\ 
\bottomrule
\end{tabular}
}
\label{tab:evaluation}
\end{table*}
\subsection{Evaluation of Dual-Metrics in Assessing GPT's Writing Capacity}

\subsubsection{Content Quality.}

All of writings with both of writing strategies are assessed by three AI grading platforms: Study Fetch, QuillBot, and Grammarly. The writing qualities such as grammar, fluency, clarity, relevance, organization, style, source formatting, and adherence to academic conventions are assessed. The full score is 100. 

On average, the writing score across three topics using two different writing methods was quite similar, the performance of all of writing strategies is over 80, with a maximum difference of only 5.34. The highest score, 86.67, was given for the GPT-assisted writing on the topic of \textit{Teacher Agency} and \textit{Student Agency}, while the lowest score, 81.33, was for the GPT-only version on the topic of \textit{Student Agency}. The detailed performance comparisons of grading software, along with the average scores for each proposal across different topics, are presented in \Cref{tab:evaluation}).

\subsubsection{Validity of References.}

Regarding the accuracy of references in the original proposals generated by LLMs, no clear pattern emerges in the error rates across the two writing methods and three topics (see \Cref{Percentage of correct references-1}). The lowest correctness rate (38.89\%) is observed in the \textit{Teacher Agency} proposal written with GPT-assisted. In contrast, the highest correctness rate—100\% accuracy with no citation errors—is found in the \textit{Teacher Identity} proposals generated both using GPT-only and GPT-assisted writing approaches.


\begin{table}[]
\centering
\caption{Percentage of correct references in each proposal across different writing strategies before applying iterative prompting method (SA: Student Agency, TA: Teacher Agency, TI: Teacher Identity)}
\label{Percentage of correct references-1}
\resizebox{0.95\linewidth}{!}{ 
\begin{tabular}{c|ccc|ccc}
\hline
& \multicolumn{3}{c|}{GPT-Only} & \multicolumn{3}{c}{GPT-Assisted} \\ \hline
Reference Criteria & SA & TA & TI & SA & TA & TI \\ \hline
AI-cited References & 8 & 10 & 9 & 5 & 18 & 10 \\
Correct AI-cited References & 5 & 4 & 9 & 4 & 7 & 10 \\
Correctness Percentage & 62.5\% & 40\% & 100\% & 80\% & 38.89\% & 100\% \\ \hline
\end{tabular}
}
\end{table}

Apart from the fact-checked references in the reference list discussed above, errors in referencing were also found within the text. In addition to standard literature citations, some in-text citations contained errors, such as the inclusion of non-academic symbols or referencing an article by its file name. The file name errors occurred because each document was initially labeled with a generic term, such as \textit{``Document 1"} rather than following proper citation conventions.


\subsection{Evaluation of Iterative Prompting Method}


\subsubsection{Content Quality.}

\begin{figure}
    \centering
    \includegraphics[width=0.95\linewidth]{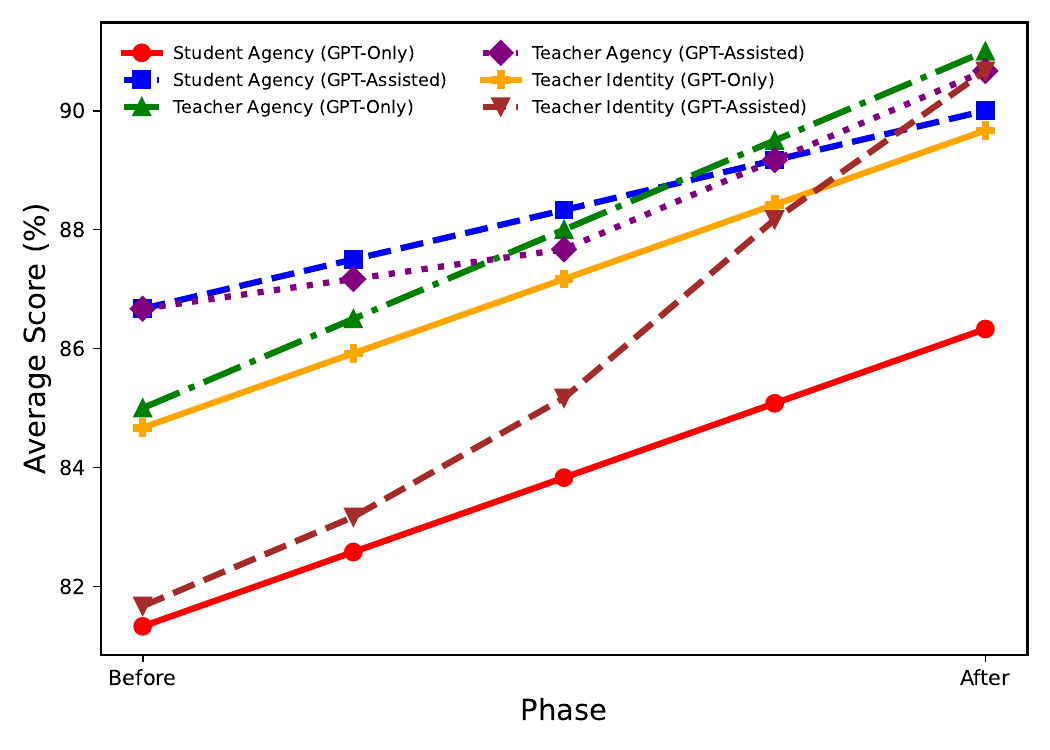}
    \caption{Content quality improvement after iterative prompting}
    \label{Content Quality Improvement After Iterative Prompting}
\end{figure}

After implementing the iterative prompting method, the content quality scores of all newly generated proposals improved and exhibited more stable performance (see \Cref{Content Quality Improvement After Iterative Prompting}). The most significant improvement was observed in GPT-assisted proposal on \textit{Teacher Identity}, with an 11.02\% increase and an average score of 90.67. In contrast, the smallest improvement occurred in GPT-assisted proposals on \textit{Student Agency}, showing a 3.84\% increase with an average score of 90. Despite variations in the degree of improvement, the results consistently indicate a clear enhancement in content quality following iterative prompting.

\subsubsection{Reference Validity.}

After implementing iterative prompting method, the introduction of reference correction instructions improved citation accuracy, with correctness rates stabilizing after the third round. Among the 6 research proposals generated using two writing strategies, two initially had no reference errors or fabrications, achieving a 100\% correctness rate without requiring correction. Additionally, one proposal remained unchanged throughout the process. For the remaining proposals, referencing accuracy consistently improved over three rounds of prompting.

\begin{figure}
    \centering
    \includegraphics[width=0.45\textwidth]{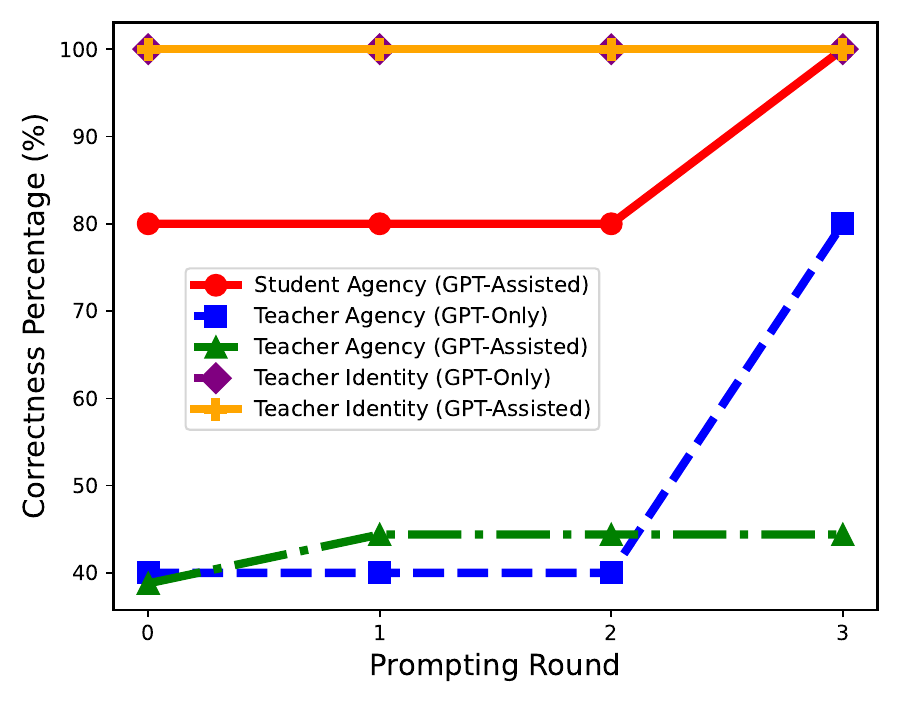}
    \caption{Percentage of correct references across different prompting rounds}
    \label{Percentage of correct references across different prompting rounds}
\end{figure}

The most substantial improvement was observed in the \textit{Teacher Agency} proposal generated by GPT-only, where reference correctness increased from 40\% to 80\%, marking a 40\% gain. Similarly, the GPT-assisted proposal on \textit{Teacher Agency} showed a moderate improvement of 5.6\%, rising from 38.8\% to 44.4\%. For GPT-assisted proposal in \textit{Student Agency}, referencing accuracy reached 100\%, up from 80\%. In contrast, proposals on \textit{Teacher Identity}, generated by both GPT-only and GPT-assisted methods, exhibited no reference errors or fabrications, maintaining 100\% correctness (see \Cref{Percentage of correct references across different prompting rounds}). Additionally, in-text citation errors, such as the use of non-academic symbols and referencing by file names, were effectively addressed through the iterative prompting method.


\noindent

\noindent
\section{SUMMARY} 
\label{Summary}

In this study, we employ ChatGPT-4o to generate academically sound, high-quality research proposals. To evaluate the writing capabilities and potential of LLMs, we adopt both standard GPT-only and GPT-assisted writing approaches. To effectively assess the writing capabilities of LLMs, we introduce two key evaluation metrics: content quality and reference validity. Additionally, we implement an iterative prompting method aimed at enhancing content quality and reducing inaccuracies and fabrications in references generated by LLMs. Our results show that the dual-metrics evaluation rigorously quantifies ChatGPT's writing capabilities, while iterative prompting enhances content quality, reduces errors, and addresses ethical concerns in reference generation. This proposal writing, evaluation, and improvement framework offers users a practical way to generate high-quality research proposals tailored to their needs. Future research can build upon this work by developing more efficient writing strategies and advanced methods to further enhance the writing capabilities of LLMs.

\def\refname{REFERENCES}

\bibliographystyle{IEEEtran} 
\bibliography{sample-base}

\begin{IEEEbiography}{Jing Ren}{\,} is a PhD student in Education from University of Technology Sydney. Contact her at jing.ren-2@student.uts.edu.au 
\end{IEEEbiography}

\begin{IEEEbiography}{Weiqi Wang}{\,} (Member, IEEE) is currently a postdoctoral researcher in the Faculty of Engineering and Information Technology, University of Technology Sydney, Australia. His research interests are in security and privacy of machine learning. Contact him at weiqi.wang@uts.edu.au.\vspace*{8pt}
\end{IEEEbiography}

\clearpage
\newpage

\appendix
\renewcommand{\thefigure}{A.\arabic{figure}}
\setcounter{figure}{0} 
\section{Appendix}


\begin{figure} [H]

    \centering
    \includegraphics[width=0.45\textwidth]{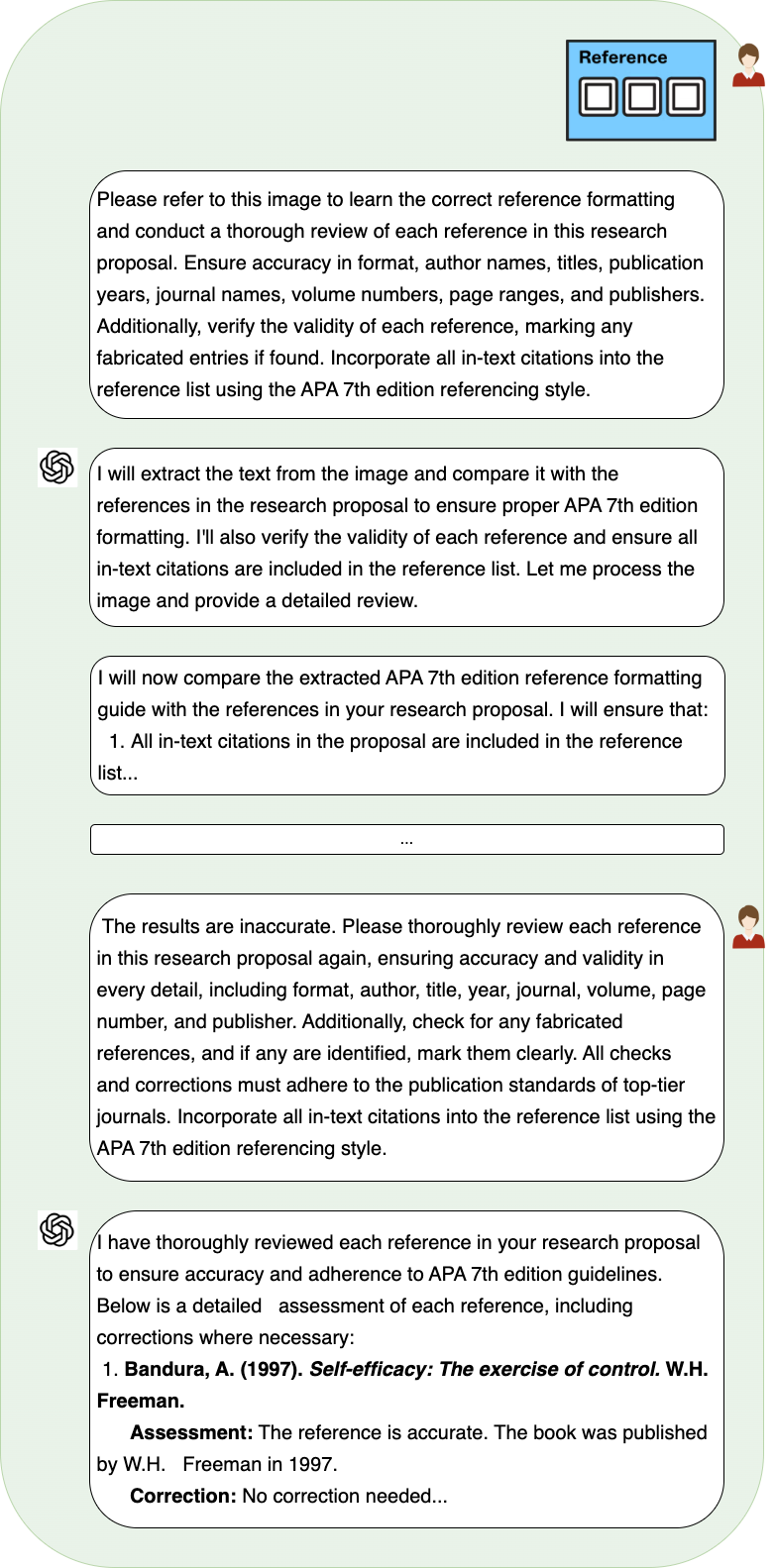}
    \caption{Improving reference accuracy in GPT-only \textit{Student Agency }proposals via iterative prompting}
    \label{Appendix: Improving Reference Accuracy}
    
\end{figure}

\subsection{Queries used in GPT-only writing}\label{appendix:gpt_only}

\begin{quote}
    You are a PhD-level researcher with significant experience in developing comprehensive research proposals. Write a 2,000-word research proposal on the topic: (\textbf{\textit{Proposal Topic}}). The proposal should be detailed and written at the level of an educational researcher, with clear and well-organized sections. Include specific and relevant subsections where appropriate, and ensure each section flows logically. Include citations from primary sources where necessary. Please use the  7th edition of APA reference style.
\end{quote}

\subsection{Queries used in GPT-assisted writing}\label{appendix:gpt_assisted}

\begin{quote}
\textbf{Query 1:} I need assistance with writing a research proposal on the \textbf{\textit{Proposal Topic}}. Can I upload 10 documents via AskYourPDF for use as references in the writing process?\\
\textbf{Query 2:} I will upload each document one at a time to facilitate better processing of the information. After uploading all documents, I will request that you write the research proposal. Please ensure you use APA 7th edition referencing style. Are you ready, or do you have any questions? \\

\textbf{Proceed to upload each document individually.}\\

\textbf{Query 3:} That was the last document. I have now provided you with 10 documents (Document 1, Document 2, Document 3, Document 4, Document 5, Document 6, Document 7, Document 8, Document 9, Document 10). Please write a 2,000-word research proposal that explores \textbf{\textit{Research Topic}}. While the discussion can include information beyond this topic, please use it as the framework for your writing. The proposal should be at a doctoral level, with in-text citations where necessary. If multiple documents support a point, use multiple citations at the end of a sentence. Ensure each section is clearly defined, well-organized, and flows logically. Please use APA 7th edition in-text citation style.
    
\end{quote}

\end{document}